\documentclass[nohyperref]{article}

\usepackage[accepted]{icml2022}

\usepackage{xcolor}
\usepackage{graphicx}
\usepackage{multirow}
\usepackage{booktabs}
\usepackage{balance}
\usepackage{subcaption}
\usepackage{setspace}
\usepackage{pifont}
\usepackage{graphics}
\usepackage{svg}
\usepackage{dblfloatfix}
\usepackage{float}
\usepackage{amsmath}
\usepackage{titlecaps}
\usepackage{algorithmic}
\allowdisplaybreaks

\DeclareMathOperator*{\argmax}{arg\,max}

\graphicspath{{images/}}

\definecolor{tblue}{RGB}{93, 142, 150}
\definecolor{tred}{RGB}{191, 97, 106}
\definecolor{dlblue}{RGB}{216, 235, 255}
\definecolor{dgreen}{RGB}{124, 155, 127}
\definecolor{dpink}{RGB}{207, 166, 208}
\definecolor{dyellow}{RGB}{255, 248, 199}
\definecolor{dgray}{RGB}{46, 49, 49}

\usepackage{amsfonts}

\icmltitlerunning{Finding Short Signals in Long Irregular Time Series with Continuous-Time Attention Policy Networks}

\begin{document}

\twocolumn[
\icmltitle{Finding Short Signals in Long Irregular Time Series with Continuous-Time Attention Policy Networks}

\icmlsetsymbol{equal}{*}

\begin{icmlauthorlist}
\icmlauthor{Thomas Hartvigsen}{mit}
\icmlauthor{Jidapa Thadajarassiri}{wpi}
\icmlauthor{Xiangnan Kong}{wpi}
\icmlauthor{Elke Rundensteiner}{wpi}
\end{icmlauthorlist}

\icmlaffiliation{mit}{MIT}
\icmlaffiliation{wpi}{WPI}

\icmlcorrespondingauthor{Thomas Hartvigsen}{tomh@mit.edu}

\icmlkeywords{}

\vskip 0.3in
]

\printAffiliationsAndNotice{}

\begin{abstract}
Irregularly-sampled time series (ITS) are native to high-impact domains like healthcare, where measurements are collected over time at uneven intervals. However, for many classification problems, only small portions of long time series are often relevant to the class label. In this case, existing ITS models often fail to classify long series since they rely on careful imputation, which easily over- or under-samples the relevant regions. Using this insight, we then propose CAT, a model that classifies multivariate ITS by explicitly seeking highly-relevant portions of an input series' timeline. CAT achieves this by integrating three components: (1) A \textit{Moment Network} learns to seek relevant moments in an ITS's continuous timeline using reinforcement learning. (2) A \textit{Receptor Network} models the temporal dynamics of both observations \textit{and} their \textit{timing} localized around predicted moments. (3) A recurrent Transition Model models the sequence of transitions between these moments, cultivating a representation with which the series is classified. Using synthetic and real data, we find that CAT outperforms ten state-of-the-art methods by finding short signals in long irregular time series.

\end{abstract}

\section{Introduction}
\label{sec:intro}
\textbf{Background.}
Irregularly-sampled time series (ITS) have uneven spaces between their observations and are common in impactful domains like healthcare \cite{hong2020holmes}, environmental science \cite{cao2018brits}, and human activity recognition \cite{singh2019multi}.
Uneven gaps can arise from many sources.
For example, in physiological streams, clinicians drive the collection of medical record data by requesting different lab tests and measurements in real time as they investigate the root causes of their patient's conditions \cite{lipton2016directly}.
\textit{Which} measurements are taken \textit{when} differs between patients.
When classifying such time series, there are often relationships between \textit{when} observations are made and the class label for the resulting time series.
For instance, sicker patients may have more measurements.
ITS can also be quite long, while the regions most-relevant to the classification may be quite short, taking up only a small portion of the timeline and creating a small \textit{signal-to-noise} ratio.
A successful model must find the best regions in the timeline at which to capture signals in both the values themselves and the patterns in \textit{when} observations were made, or \textit{informative irregularity}, while ignoring irrelevant regions.

\begin{figure}[tp]
    \centering
    \vspace{1mm}
    \includegraphics[width=0.48\textwidth]{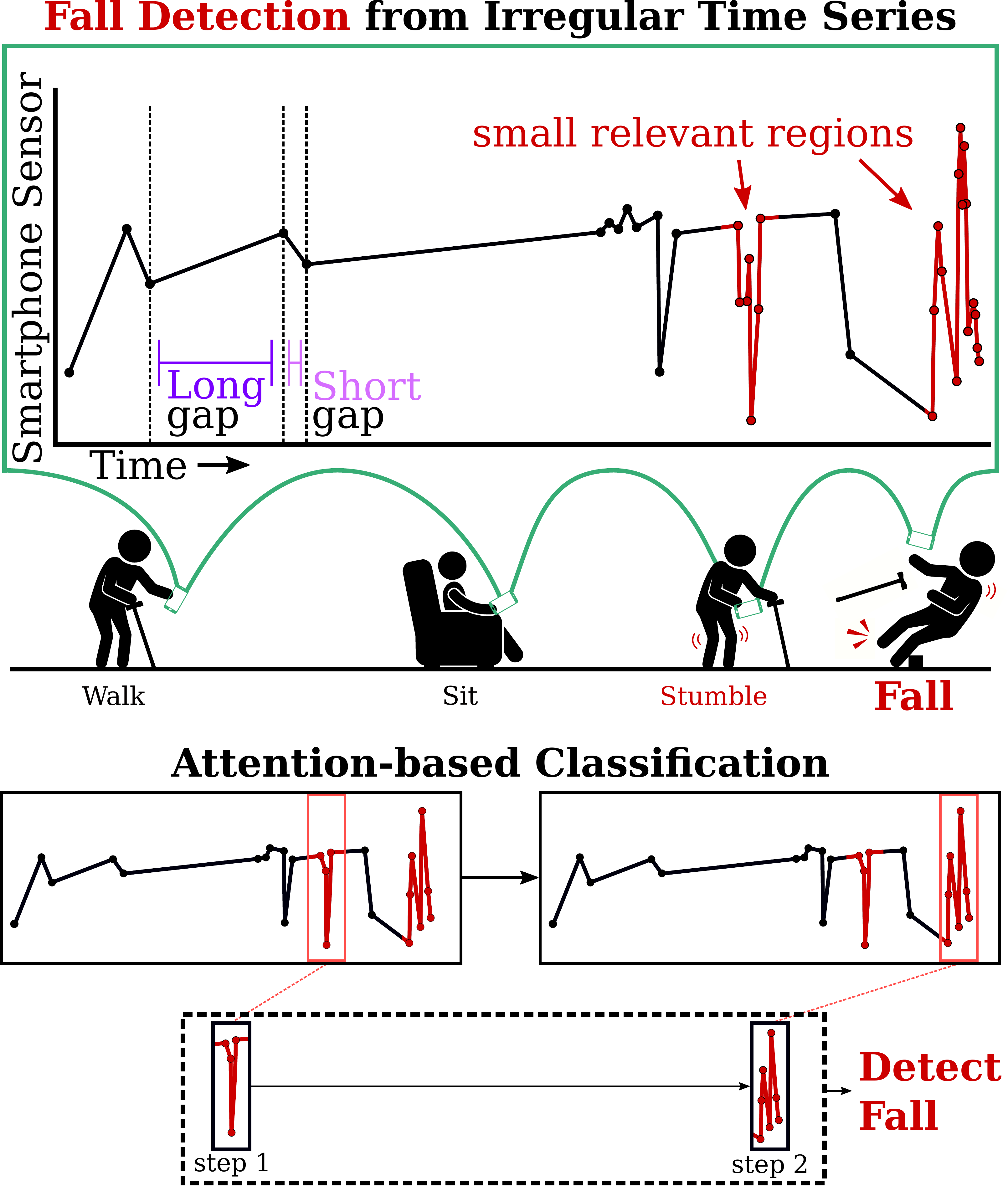}
    \vspace{-2mm}
    \caption{Attention-based classification for ITS.}
    \vspace{-5mm}
    \label{fig:prob_def}
\end{figure}

\textbf{Motivating Example.}
Consider detecting if a person \textit{Fell} using their smartphone's sensors, as illustrated in Figure \ref{fig:prob_def}.
To extend battery life, a listening probe is used to only collect data when certain conditions are met, for example when the accelerometer changes rapidly.
Since the phone is not always moving, the stored time series are naturally irregularly-sampled.
To detect a fall, some regions of the accelerometer's records are far more relevant than others.
Leading up to a fall, for instance, a person may have stumbled earlier in the day.
However, there can also be many false positives where the phone moves quickly even though the person is not falling (setting the phone down, for instance).
Additionally, \textit{when} observations are made can also be useful: if the phone moves after a long gap, the person may be getting out of bed.
Since only some regions are relevant, all a classifier needs are the few most relevant moments in the timeline.
Finding these moments is especially important for the long series that naturally exist in many domains. 


\textbf{State-of-the-art.}
There have been many recent advances in classifying ITS data, though most focus on sparse series with few observations.
Many works treat ITS classification as a \textit{missing value imputation} problem \cite{che2018recurrent,lipton2016directly,zheng2017resolving}, converting ITS to regular series then performing standard classification.
However, to capture short signals, many values need to be imputed to avoid aggregating intricate signals.
Plus, this increases the length of the series and imputes values in irrelevant regions of the timeline.
As many ITS methods rely on Recurrent Neural Networks, making time series longer will likely decay performance.
On the other hand, imputing too few values easily bypasses short signals, aggregating away crucial information.
Some works capture informative irregularity by computing statistical features such as \textit{missingness indicators} \cite{lipton2016directly} or the time since last observations \cite{che2018recurrent} as additional input variables, inflating the feature space.

Some recent works have leaned into learning continuous-time representations \textit{directly} from raw ITS data \cite{kidger2020generalised,shukla2019interpolation,li2016scalable,cheng2020learning,rubanova2019latent,de2019gru,oh2018learning,shukla2021multi}.
However, they still rely on hand-picking new \textit{reference} timesteps at which to estimate values or compute representations, falling prey to the same challenges of imputation.
To-date, these methods do not adapt their reference timesteps to the inputs. Overall, current machine learning methods for classifying ITS data are expected to underperform on long series where the relevant signals are proportionally short.

\textbf{Problem Definition.}
We specifically address the problem of \textit{Attention-based ITS Classification} (ABC), which is to classify long ITS by finding small discriminative signals in the continuous timeline, as illustrated in Figure \ref{fig:prob_def}.
Given a set of labeled ITS, 
where each series consists of one sequence of (\textit{timestep}, \textit{value}) pairs per variable, our aim is to produce a classifier that can correctly assign class labels $y$ to previously-unseen instances.
For long series, the relevant time window, or the proportion of the timeline needed for classification, may be very small in practice.
A successful model should explicitly find these \textit{discriminative moments} with which it can make an accurate classification.

\textbf{Challenges.}
Solving the ABC problem is challenging for the three following reasons:
\begin{itemize}
    \item \textit{Finding Short Signals.} Short, relevant windows of a continuous timeline can be hard to identify, akin to finding a needle in a haystack. For long series, this means that much of the timeline contains effectively irrelevant information, which a model must learn to ignore. Meanwhile, the model must also while avoiding learning spurious correlations found outside relevant regions.
    
    \item \textit{Unknown Signal Locations.} Relevant signals may occur anywhere in the continuous timeline. However, rarely are the \textit{true} signal locations labeled, so we assume no prior knowledge of which moments \textit{should} be used for classification. Still, a good model must successfully find these discriminative moments, even without supervision. 
    \item \textit{Informative Irregularity.} Discriminative information often arises in the patterns of \textit{when} observations are made \cite{rubin1976inference}. For instance, rapid measurements may indicate a sicker patient. Learning from such irregularity is often crucial to accurate classification, yet few methods exist for capturing such signals.
\end{itemize}

\textbf{Proposed Method.}
To address these challenges, we propose the \textbf{C}ontinuous-time \textbf{At}tention policy network (CAT) as an effective approach to the ABC problem.
CAT searches for relevant regions of input series via a reinforcement learning-based \textit{Moment Network}, which learns to find \textit{moments of interest} in the continuous timeline, one by one.
At each predicted moment, a \textit{Receptor Network} reads and represents the local temporal dynamics in the measurements along with patterns that exist in the timing of observations through a continuous-time density function.
Along the way, a recurrent \textit{Transition Model} constructs a discriminative representation of the \textit{transitions between moments of interest}, which is ultimately used to classify the series. 
CAT thus presents a novel paradigm for classifying ITS where intricate signals in long series are explicitly sought out and captured.
Additionally, CAT generalizes recent ITS classifiers with its flexible \textit{Receptor Network}, which can easily be augmented to leverage components of other recent ITS models.

\textbf{Contributions.}
Our contributions are as follows:
\begin{itemize}
    \item We identify a new, real problem setting for classifying irregularly-sampled time series on which existing state-of-the-art methods underperform.
    \item Using insights from this problem, we develop CAT, a novel framework for classifying long irregular time series by finding relevant moments in the \textit{continuous} timeline, generalizing recent work.
    \item We show that CAT successfully discovers intricate signals in ITS, outperforming the the main competitors on both synthetic and real-world data.
\end{itemize}

\section{Related Work}
\label{sec:related_works}
The ABC problem for ITS relates to both \textit{ITS Classification} and \textit{Input Attention}.

\textbf{Classifying Irregularly-Sampled Time Series.}
Classifying irregularly-sampled time series has recently become a popular and impactful problem as it generalizes many prior classification settings.
To-date, most approaches  \cite{lipton2016directly,zheng2017resolving,che2018recurrent} treat ITS classification as a \textit{missing value imputation} problem: Create a set of evenly-spaced bins, then aggregate multiple values within each bin and estimate one value per empty bin.
After imputation, regular time series classification may be performed. 
Some recent ITS classifiers extend beyond simple imputation options (\textit{e.g.}, mean) approaches by either including auxiliary information such as a \textit{missingness-indicator} \cite{lipton2016directly} or \textit{time-since-last-observation} \cite{che2018recurrent} as extra features to preserve properties found in the irregularity.
Others build more complex value estimators by either learning generative models \cite{cheng2020learning}, using differentiable gaussian kernel adapters \cite{shukla2019interpolation}, or including decay mechanisms in Recurrent Neural Networks (RNN) to encode information-loss when variables go unobserved over long periods of time \cite{mozer2017discrete,che2018recurrent}.
Many recent works have also begun parameterizing differential equations to serve as time series models \cite{kidger2020neural,lechner2020learning,rubanova2019latent,jia2019neural,hasani2021liquid,schirmer2022modeling,salvi2022neural}, though most still estimate values at hand-picked time steps, then use the estimated values for classification.

Some recent models have also integrated attention mechanisms into ITS classification \cite{shukla2021multi,chen2021continuous,tan2021cooperative}.
However, they still hand-pick reference timesteps for each input time series. 
Given long ITS with short signals, this decision is hugely impactful, as we show in our experiments.
Moreover, by relying on RNNs for classification, these recent methods easily fail to capture signals when the number of estimated values gets too large.
This requires the RNN to filter out many irrelevant timesteps in a long series, which is notoriously challenging due to both their slow inference and vanishing gradients \cite{hochreiter1998vanishing}.

\textbf{Input Attention.}
The goal of \textit{Input Attention} is to discover relevant regions in the \textit{input} space of a given instance and it has recently broken major ground in classifying images \cite{mnih2014recurrent}, graphs \cite{lee2019attention}, text \cite{sood2020improving}, and regularly-spaced time series \cite{ismail2019input}.
We refer to this as \textit{input} attention, as such methods search for relevant regions in the \textit{input space} of each instance.
This approach is particularly impactful when inputs are high-dimensional as it explicitly disregards irrelevant regions of the input space. 
These methods also aid interpretability by clearly displaying which regions of an input were used to make a classification.
Input attention has yet to be considered for ITS despite strong implications of successful models.

Input attention differs from \textit{attention mechanisms for recurrent neural networks} \cite{bahdanau2014neural}, where attention distributions are predicted over the timesteps in the \textit{latent} space of an RNN. 
While there is some conceptual overlap, input attention is more data-driven in that it finds regions in the \textit{input} space as opposed to the \textit{latent} space of this specific neural network architecture.


\section{Methodology}
\label{sec:methods}
\begin{figure*}[tp]
    \centering
    \includegraphics[width=0.92\textwidth]{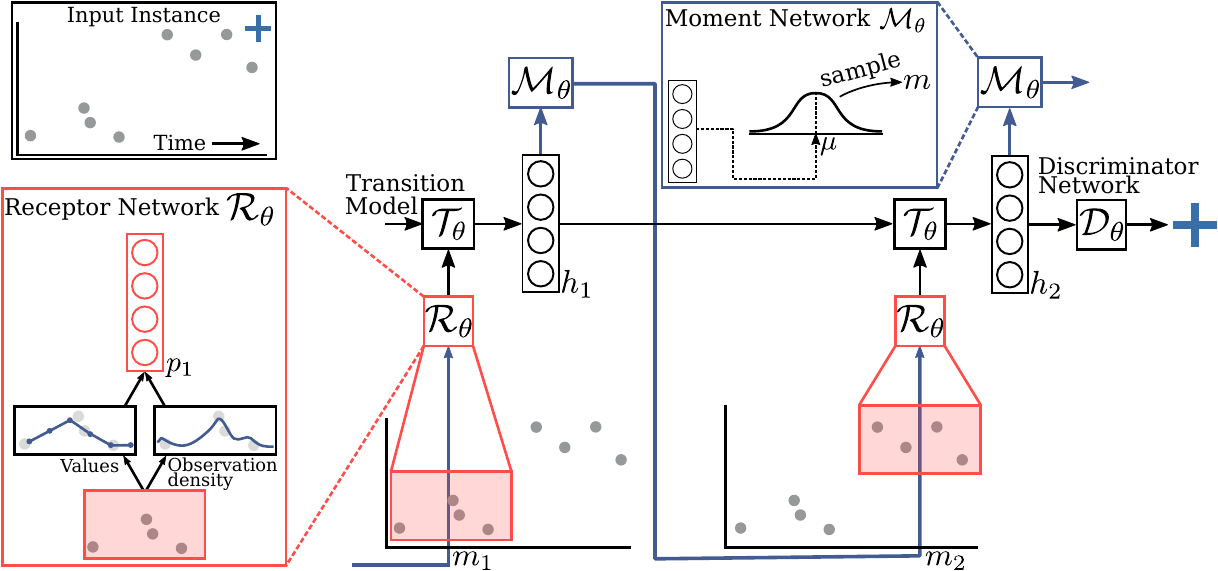}
    \caption{Overview of CAT. The \textit{Receptor Network} models input values and irregularity around a moment $m_i$ in the continuous timeline: $\hat{x}_i=\mathcal{R}(X, m)$. The \textit{Transition Model} then updates its hidden state $h_i = \mathcal{T}(\hat{x}_i)$, modeling the transitions between \textit{moments}. Then, the \textit{Moment Network} parameterizes a Normal distribution from which it samples the next moment $m_{i+1} = \mathcal{M}(h_i)$. After iterating $K$ times, the \textit{Discriminator Network} predicts the final class: $y=\mathcal{D}(h_K)$, classifying the entire series.}
    \label{fig:architecture}
\end{figure*}


\subsection{Problem Formulation}
Given a set of $N$ labeled irregularly-sampled time series $\mathcal{D} = \{(X_i, y_i)\}_{i=0}^N$, consider the $D$ variables of instance $X_i = [X_i^1, \dots, X_i^D]$.
To aid readability, all descriptions are provided
in terms of one instance and one variable wherever possible.
For each variable $d$, $X^d = [(t^d_1, v^d_1), \dots, (t^d_{T^d}, v^d_{T^d})]$, where $t^d_i$ is the $i$-th timestamp of the $d$-th variable and $v^d_i$ is its corresponding value.
Timestamps $t$ may differ between variables and the number of observations $T^d$ may be unique to variable $d$.
We also assume that the inputs $X$ have short signals: Most of the relevant information comes from a small proportion of a series' timeline.
There may still be multiple relevant regions, however.
The goal is to learn a function $f:\mathbb{X} \to \mathcal{Y}$ that accurately maps input $X \in \mathbb{X}$ to its class $y \in \mathcal{Y}$ for previously-unseen time series, where $\mathbb{X}$ is the input space of ITS and $\mathcal{Y} =\{0, \dots, C\}$ is the set of $C$ classes.

\subsection{Proposed Method}

We propose a \textbf{C}ontinuous-time \textbf{At}tention Policy Network (CAT), a novel model containing four key steps that work in concert to find short discriminative signals in long ITS:
\begin{enumerate}
    \item A \textit{Receptor Network} learns to model ITS observations (both the raw values \textit{and} informative irregularity) local to a given \textit{moment of interest}.
    \item A recurrent \textit{Transition Model} represents the Receptor Network's findings across multiple moments.
    \item A reinforcement learning \textit{Moment Network} predicts \textit{moments of interest} based on the Transition Model. A \textit{moment of interest} is timestamp around which relevant information may exist.
    \item After $k$ repetitions of Steps 1-3, a \textit{Discriminator Network} classifies $X$ using all steps.
\end{enumerate}


Beyond state-of-the-art performance, a clear benefit of CAT is its novel framework for classifying time series:
CAT decomposes a time series into a sequence of local representations that is discriminative in (1) which subsequences are modeled, and (2) their relative order.
This approach is more flexible than rigidly reading a time series from either left-to-right, right-to-left, or all-at-once; our model adapts the processing order to the input.
Since ordering is discrete, Reinforcement Learning is a natural fit. That is, we let the model pick the order, then reward or penalize based on the final classification.
This technical novelty helps CAT stand out from alternative ITS models.



\subsubsection{Receptor Network}
First, the Receptor Network  $\mathcal{R}_\theta$ creates a vector representation of values and irregularity \textit{local} to a given moment of interest $m_i$.
So given $m_i \in [0, \max T]$, $\mathcal{R}_\theta$ predicts a vector $\hat{\mathbf{x}}_i$, representing the local \textit{values} and \textit{informative irregularity} within a width-$\delta$ window of $X$ centered on moment $m_i$, where $\max T$ is the largest timestamp in $X$.
Thus $\mathcal{R}_\theta$ can be placed anywhere in the continuous timeline, where it will proceed to model local signals.
In our experiments, the first moment, $m_0$, is sampled from a uniform distribution across the timeline.
To compute local representations of both values \textit{and} irregularity, we compute two $w$-dimensional vectors per variable: $\mathbf{p}$ represents $X$'s values and $\mathbf{q}$ represents informative irregularity, which are then encoded into a shared representation $\hat{\mathbf{x}}$.
For readability, we describe $\mathcal{R}$ for one variable, omitting superscripts $d$ since all variables are processed the same way and in parallel.

To compute vector $\hat{\mathbf{x}}$, all timestamps and values within this window are first extracted into two vectors: $\tau$ is a sequence of timestamps in the window $[m_i-\frac{\delta}{2}, m_i+\frac{\delta}{2}]$, and $\nu$ contains their corresponding values.
We use $[\ ]$ to denote a range in the timeline beginning at the real time $m_i-\frac{\delta}{2}$ and ending at time $m_i+\frac{\delta}{2}$.

To compute $\mathbf{p}$, the representation of the values within the window surrounding $m_i$, we linearly interpolate values $\nu$ to estimate $w$ values at a set of new timestamps. 
The $j$-th element of $\mathbf{p}$ can be interpolated with respect a to timestamp $t^\prime = m_i-\frac{\delta}{2}+\frac{j\delta}{w}$ for $j=\{1, \dots, w\}$ as
\begin{equation*}
    \mathbf{p}_j = \frac{\left(\text{SG}(t^\prime, \tau)-t^\prime\right)\nu_{\text{LL}(t^\prime, \tau)} + \left(t^\prime - \text{LL}(t^\prime, \tau)\right)\nu_{\text{SG}(t^\prime, \tau)}  }{\text{SG}(t^\prime, \tau)-\text{LL}(t^\prime, \tau)},
\end{equation*}
where $\text{LL}(t^\prime, \tau)$ is the largest timestamp less than $t^\prime$ and $\nu_{\text{LL}(t^\prime, \tau)}$ is its corresponding value.
Similarly, $\text{SG}(t^\prime, \tau)$ is the smallest timestamp greater than $t^\prime$ and $\nu_{\text{SG}(t^\prime, \tau)}$ is its corresponding value.
By iterating $j$ across integers from 1 to $w$, we compute $w$ evenly-spaced values representing the local observations.
If a timestamp $t^\prime > \max T$ or $t^\prime < \min T$, the nearest value in the window is returned, flattening the edges of the window.
If no observations occur in the window, we set $\hat{x} = \{0\}^w$.

To compute $\mathbf{q}$, which represents \textit{informative irregularity} within the window, we learn a function to represent the \textit{timing} of observations, quantifying irregularity through the squared exponential kernel, inspired by \cite{li2016scalable}.
Thus the $j$-th element of $\mathbf{q}$ as computed with respect to each $t^\prime = m_i-\frac{\delta}{2}+\frac{j\delta}{w}$ for $j=\{1, \dots, w\}$ is
\begin{equation}\label{eqn:density}
    \mathbf{q}_j = \sum_{k=1}^{|\tau|} e^{-\alpha(t^\prime-\tau_K)},
\end{equation}
\noindent where $\tau_K$ is the $K$-th element of sequence $\tau$. 
Thus the \textit{timing} of the observations is converted to a sequence of densities, which often change by class \cite{lipton2016directly}.
$\alpha$ controls the kernel distance between $t^\prime$ and $\tau_K$ and can be picked or learned during training \cite{shukla2019interpolation}.

Since the output of the Receptor Network will eventually be used by the Moment Network to predict the next moment $m_{i+1}$, we also compute $\mathbf{p}$ and $\mathbf{q}$ for each variable at two granularities: One for fine-grained local information, one for coarse-grained representation of the entire series that is useful for both capturing long-term trends and for finding the next moments of interest, inspired by \cite{mnih2014recurrent}.
After computing $\mathbf{p}$ and $\mathbf{q}$, a neural network predicts a $L$-dimensional representation $\mathbf{\hat{x}}_i$, creating a dense, vector representation of the width-$\delta$ window surrounding moment $m_i$:
\begin{equation}
    \mathbf{\hat{x}}_i = \psi(\mathbf{W}[\text{F}(\{\mathbf{p}^d\}_{d=1}^D), \text{F}(\{\mathbf{q}^d\}_{d=1}^D)] + \mathbf{b}),
\end{equation}
where F$(\cdot)$ and $[\cdot]$ denote flattening and concatenation, respectively. $\mathbf{W}$ and $\mathbf{b}$ are a matrix and vector of learnable parameters of shape $L \times 4w$ and $4w$. $\psi$ is the rectified linear unit.
To also incorporate \textit{where} the collected data come from in the timeline, we concatenate $m_i$ with $\mathbf{\hat{x}}_i$ before passing it to the Transition Model.


\subsubsection{Transition Model}
Next, the Transition Model $\mathcal{T}_\theta$ represents the \textit{transitions} between information gathered at each moment of interest.
We follow the state-of-the-art for a vast array of sequential learning tasks and implement this component as an RNN, creating one $H$-dimensional vector representation $\mathbf{h}_i$ per moment-of-interest.
To avoid vanishing gradients, 
we use a Gated Recurrent Unit (GRU) \cite{cho2014learning} to compute the hidden state $\mathbf{h}_i$.

This recurrent component takes only $K$ steps and $K$ is typically kept very low ($K=3$ in our experiments).
In contrast, most recent models instead step through a large number of imputed timestamps $T$ (typically $T \gg K$) creating slow models that are hard to optimize.

\subsubsection{Moment Network}
Next, the \textit{Moment Network} $\mathcal{M}_\theta$ uses the hidden state $h_i$ and predicts the next moment-of-interest $m_{i+1}$.
There are no ground-truth moments, so we frame this component as a Partially-Observable Markov Decision Process (POMDP), similar to \cite{mnih2014recurrent}.
We follow the standard approach and solve this POMDP using on-policy reinforcement learning.
In this way, the hidden state $\mathbf{h}_i$ from the \textit{Transition Model} serves as an observation from the environment (representing the data collected at all prior moments of interest).
The possible actions include all real-valued timestamps between 0 and $\max T$, and we define the reward to be the final classification success.
The goal is to learn a policy $\pi(\mathbf{h}_i)$ that predicts the next moment $m_{i+1}$.

Since there are infinitely-many moments in the continuous timeline, we parameterize the mean $\mu_i$ of a Normal distribution with fixed variance from which we \textit{sample} real-valued $m_{i+1}$.
To acquire good samples, $\mathbf{h}_i$ is first projected into a one-dimensional probabilistic space by a neural network: $\mu_i = \sigma\left(\mathbf{W}\mathbf{h}_i+\mathbf{b})\right).$ 
We then scale $\mu_i$ by multiplying with $\max T$ and sample moment $m_{i+1} \sim \mathcal{N}(\mu_i, \sigma)$, with tunable $\sigma$.
If $m_{i+1} > \max T$, we re-assign $m_{i+1} := \max T$, and if $m_{i+1} < 0$, $m_{i+1}:=0$.
To train $\mathcal{M}_\theta$, we set reward $r_i = 1$ if the final classification is accurate, otherwise $r_i = -1$.
The Moment Network thus seeks \textit{discriminative} regions, which lead to the highest rewards. 

CAT predicts $K$ moments of interest, iteratively cycling between the Receptor Network, Moment Network, and Transition Model $K$ times.
This packs information from $K$ steps into the final hidden state $\mathbf{h}_K$.

\subsubsection{Discriminator Network}
The final component of CAT is a \textit{Discriminator Network} $\mathcal{D}_\theta$, which learns to project the Transition Model's final hidden state $\mathbf{h}_K$ into a $C$-dimensional probabilistic space in which it predicts $\hat{y}$ to be $X$'s class label.
This final classification is made via a single linear layer:
    $\hat{y} = \text{softmax}(\mathbf{W}\mathbf{h}_K+\mathbf{b}).$
The discriminator is naturally connected to the transition model, so is easily expandable according to the required complexity of a task.

\subsubsection{CAT Training}\label{sec:optimization}
The Receptor Network, Transition Model, and Discriminator are optimized together to predict $\hat{y}$ accurately by minimizing cross entropy: 
\begin{align}\label{eqn:cross_entropy}
    \mathcal{L}_{\text{s}}(\theta_s) = -\sum_{c=0}^C y_c \log\hat{y}_c,
\end{align}
where $y_c$ is 1 if $X$ is in class $c$ and $\hat{y}_c$ is the corresponding prediction.
$\theta_s$ denotes these networks' parameters.

The Moment Network, on the other hand, samples the moments, so its learning objective is the maximization of the expected reward: $R = \sum_{i=0}^K r_i$, so $\theta_{\text{rl}}^* = \argmax_{\theta_{\text{rl}}}\mathbb{E}[R]$,
where $\theta_{\text{rl}}^*$ is the optimal parameters for the \textit{Moment Network}.
However, this is not differentiable.

To maximize $\mathbb{E}[R]$ using backpropagation, we follow the standard protocol for on-policy reinforcement learning and optimize the Moment Network's policy using the REINFORCE algorithm \cite{williams1992simple}.
Thus, we use a well-justified surrogate loss function that \textit{is} differentiable, allowing for optimization by taking steps in the direction of $\mathbb{E}[\nabla\log\pi(h_{0:k}, \mu_{0:k},r_{0:k})R]$.
Thus the gradient can then be approximated for the predicted moments. 
Thus learning progresses, but there may be high variance in the policy updates since this is not the \textit{true} gradient for maximizing $\mathbb{E}[R]$.
To reduce variance, we employ the commonly-used \textit{baseline} approach to approximate the expected reward, with which we may adjust the raw reward values, as shown in Equation \ref{eqn:reinforce}. 
Here, $b_j$ is a baseline predicted by a two-layer neural network and its predictions approximate the mean $R$ by reducing the mean squared error between $b_j$ and the average $R$.
The weights $\theta_\text{rl}$ are thus updated by how much better than average are the outcomes.
\begin{align}
    \mathcal{L}_{\text{rl}}(\theta_\text{rl}) = -\mathbb{E}\Bigg[\sum_{i=0}^{k} \log \pi(m_i|h_i)\bigg[\sum_{j=i}^{k} \big(R - b_{j}\big)\bigg]\Bigg]\label{eqn:reinforce}
\end{align}

Finally, the entire network can be optimized jointly via gradient descent on the sum of Equations 3 and 4: $\mathcal{L}(\theta) = \mathcal{L}_{s}(\theta_\text{s}) + \mathcal{L}_\text{rl}(\theta_\text{rl})$, 
where $\theta$ denotes CAT's parameters.

\section{Experiments}
\label{sec:experiments}
\subsection{Datasets}\label{sec:datasets}

We evaluate CAT using one synthetic dataset and five real-world publicly-available datasets. 

\textsc{M$\Pi$}:
We develop a synthetic binary classification dataset to demonstrate that CAT indeed finds short signals in long ITS data.
To add signals for different classes, we center a width-$\Delta$ discriminative region around a random moment in the timeline for each time series.
The values for the timestamps within the width-$\Delta$ window take one of two forms, depending on the class.
One class is characterized by the values $\{1, 1, 1\}$ (``$\Pi$''-shaped), and the other by the values $\{1, 0, 1\}$ (``M''-shaped).
The timestamps corresponding to these values are evenly-spaced in the width-$\Delta$ window.
All timestamps \textit{not} in the discriminative region are sampled uniformly across the timeline and values are sampled from a Normal distribution $\mathcal{N}(0, 1)$. 
In selecting $\Delta$, we determine the signal-to-noise ratio of the data: A small $\Delta$ means that the ``$\Pi$'' or ``M'' signals happen in a short period of time, so overlooking the signal is punished more.
We generate 5000 time series, each with 500 timestamps, and have an equal number of instances for each class.

\texttt{UWave} \cite{liu2009uwave}:
The popular \texttt{UWave} dataset contains 4478 length-945 gesture pattern time series collected from a handheld device.
Each series is a member of one of eight classes.
We follow the preprocessing procedure outlined by \cite{li2016scalable}, randomly downsampling to 10\% of the original values to create irregularity.

\texttt{ExtraSensory} \cite{vaizman2017recognizing}: Following \cite{hartvigsen2022stop}, we augment existing human activity data by simulating listening probes on smartphone data.
Listening probes collect data from devices only when certain conditions are met, creating realistic ITS.
For example, consider detecting hand tremors for digital health \cite{garcia2016smartphone}.
A listening probe on a smartphone's accelerometer will collect data only when the phone moves rapidly, capturing hand tremors while the phone is carried.
However, false positives are common: when the phone is set down or dropped, data are \textit{also} collected, resulting in irrelevant regions.
Our sampling is more realistic than prior works, which randomly downsample without encoding meaning into the irregularity of samples.

We extract four disjoint, non-overlapping datasets from the challenging \texttt{ExtraSensory} human activity recognition database \cite{vaizman2017recognizing} via a simulated listening probe on the 3-dimensional (x, y, and z axes) accelerometer records.
When the norm of the difference between consecutive records surpasses a threshold $\gamma=0.001$, the corresponding accelerometer data are collected.
We collect four datasets, one for each of four human activities: \textsc{Walking} (2636 time series), \textsc{Running} (1066 time series), \textsc{LyingDown} (7426 time series), and \textsc{Sleeping} (9276 time series).
For each class, we extract data for the person who performed the activity the most since people's activity patterns are often incomparable.
We then break each series into windows of 200 timestamps, then apply the listening probe. The task is to detect whether the person performed the activity within this window.
We finally balance each dataset to have an equal number of positive and negative series and ensure no extracted segments overlap.

\subsection{Compared Methods}
We compare CAT to ten recent ITS classifiers.
The first four methods are use imputation and feature expansions: linear interpolation (GRU-interp), mean imputation (GRU-mean), mean imputation with extra time-since-last-observation features (GRU-$\Delta t$), and mean imputation with a missingness indicator (GRU-S) \cite{lipton2016directly}.
The second group contains state-of-the-art ITS classifiers: GRU-Decay \cite{mozer2017discrete}, GRU-D \cite{che2018recurrent}, IPN \cite{shukla2019interpolation}, mTAN \cite{shukla2021multi}, and NCDE \cite{kidger2020neural}.
We also ablate CAT by replacing the Moment Network with randomly-selected moments of interest, which we refer to as CAT w/o Moment.

\begin{figure}[t]
    \centering
    \includegraphics[width=0.47\textwidth]{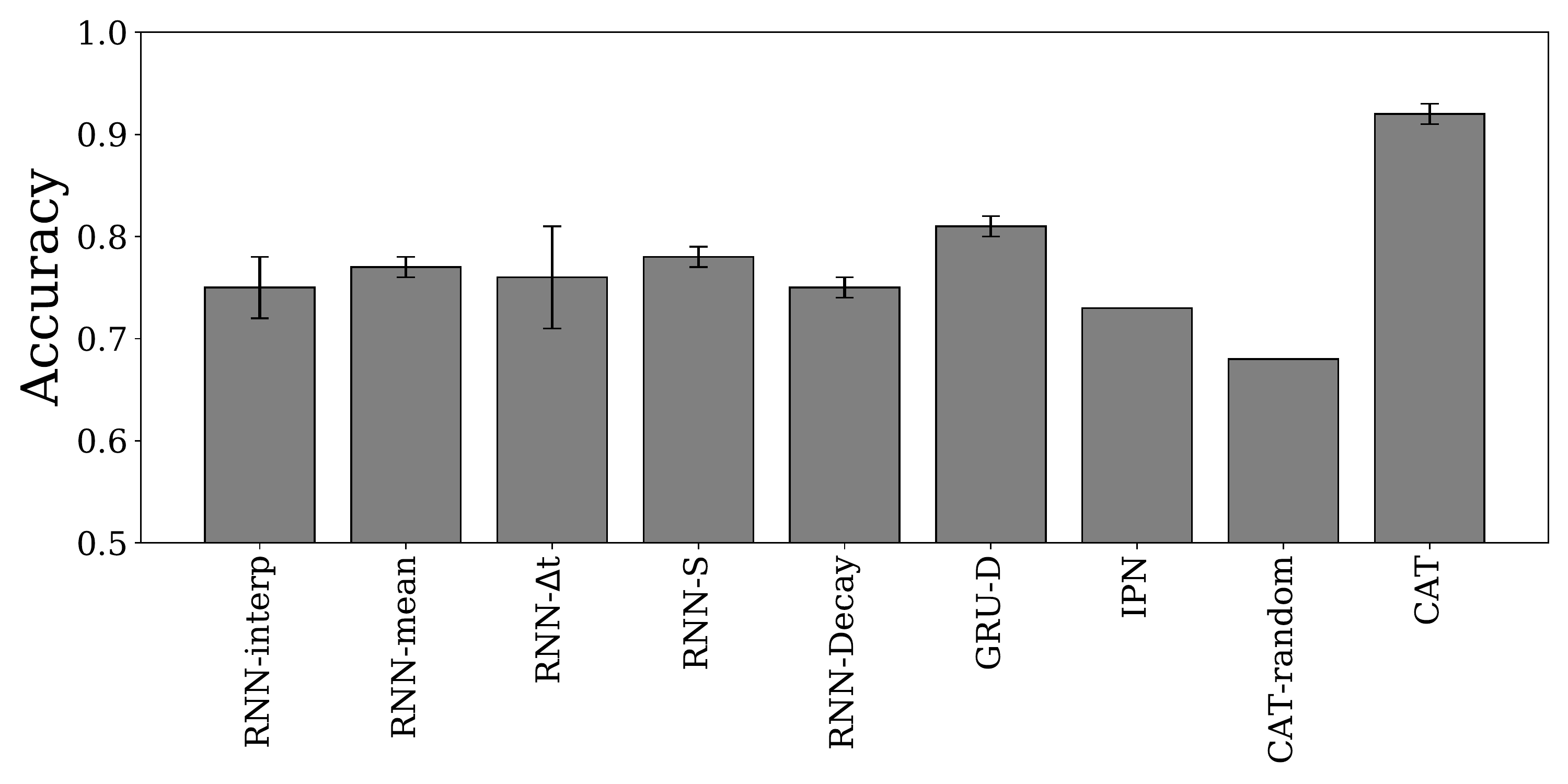}
    \vspace{-5mm}
    \caption{Multi-class classification with long \texttt{UWave} series. CAT outperforms recent methods when sampling many points from the timeline of each \texttt{UWave} time series.}
    \label{fig:uwave}
\end{figure}

\subsection{Implementation Details}
For the \texttt{UWave} dataset, we use a standard 80\% training, 10\% validation, and 10\% testing split.
The \texttt{ExtraSensory} datasets contain instances taken from different windows along a single timeline.
To avoid cross-contamination, we split instances \textit{in time}, aiming for 80\% training and 20\% testing splits.
The training/testing process is repeated five times and we report the average and standard deviation for all experiments.
All methods use 64-dimensional hidden states for their respective RNNs.
For CAT, we set $k=3$, use a 50-dimensional representation for the Receptor Network, and set $\alpha = 100$ in Equation \ref{eqn:density}.
All models are optimized using Adam 
with a learning rate of $1e^{-3}$ and weight decay of $1e^{-5}$ and all methods are run until their losses converge, taking around 200 epochs.
Each model is implemented in PyTorch in our public code.

    

\subsection{Experimental Results}



\subsubsection{Experiments on Real-World Data.}
First, we demonstrate that CAT indeed handles long series better than the state-of-the-art methods.
To achieve this, we impute the \texttt{UWave} data with 200 timestamps, which is much higher than prior experiments \cite{shukla2019interpolation}.
For ease of comparison, we also have CAT observe the data at the same ``resolution'' by setting $w = \delta*200$ where $\delta$ is the receptor-width hyperparameter.
This resolution can be tuned within CAT.
Our results are reported in Figure \ref{fig:uwave}.
As expected, CAT achieves state-of-the-art accuracy on these data while the compared methods underperform their accuracy with roughly 100 imputed values.
This indicates that CAT is far more robust to longer series than the state-of-the-art ITS classifiers.


Second, we show that CAT successfully captures \textit{informative irregularity} in long series, as indicated by our results on the human activity recognition datasets (\textsc{Walking}, \textsc{Running}, \textsc{LyingDown}, and \textsc{Sleeping}).
We compare all models using two settings: infrequent imputation (200 values) and frequent imputation (500 values).
Intuitively, \textit{frequent} imputation leads to clearer signals, as there are more values imputed on the signal, while \textit{infrequent} imputation leads to unclear signals.
To successfully classify these data given infrequent imputation, finding the relevant regions of the data is more important. 
On the other hand, frequent imputations provide clear signals but come with the added noise, requiring explicit discovery of the relevant regions.
Again, to compare with other methods, we set $w=\delta*200$ and $w=\delta*500$ for each respective frequency.

\begin{table}[t]
\centering
\caption{Accuracy with \textit{infrequently}-imputed values for the Human Activity datasets. \textbf{Bold} indicates best performance.}\label{tab:low_res}
\vspace{-2mm}
\resizebox{\linewidth}{!}{
\begin{tabular}{lcccc}
\toprule
\multirow{2.25}{*}{\textbf{Methods}} &  \multicolumn{4}{c}{\textbf{Datasets}}\\
\cmidrule{2-5}
    & \texttt{Walking}
    & \texttt{Running}
    & \texttt{Lying Down}
    & \texttt{Sleeping}\\      
\midrule
GRU-interp & 0.65 (0.02) 	 & 0.52 (0.04) 	 & 0.83 (0.05) 	 & 0.76 (0.05) \\
GRU-mean & 0.64 (0.02) 	 & 0.52 (0.09) 	 & 0.78 (0.04) 	 & 0.79 (0.01) \\
GRU-$\Delta$t & 0.64 (0.01) 	 & 0.52 (0.07) 	 & 0.83 (0.04) 	 & 0.79 (0.03) \\
GRU-S & 0.64 (0.01) 	 & 0.46 (0.03) 	 & 0.82 (0.08) 	 & 0.76 (0.03) \\
GRU-Decay & 0.64 (0.02) 	 & 0.44 (0.03) 	 & 0.77 (0.04) 	 & 0.78 (0.03) \\
GRU-D & 0.65 (0.01) 	 & 0.43 (0.01) 	 & 0.78 (0.03) 	 & 0.76 (0.03) \\
IPN & 0.65 (0.01) 	&  0.46 (0.03) 	 & 0.85 (0.04) 	 & 0.77 (0.03) \\
mTAN & 0.68 (0.02) 	&  0.56 (0.01) 	 & 0.73 (0.00) 	 & 0.76 (0.02) \\
NCDE & 0.66 (0.00) 	& 0.53 (0.00) 	& 0.75 (0.00) 	& 0.71 (0.00) \\
\midrule
CAT w/o Moment & 0.60 (0.02) 	 & 0.47 (0.01) 	 & 0.76 (0.01) 	 & 0.73 (0.01) \\
CAT (ours) & \textbf{0.81 (0.03)}	 & \textbf{0.62 (0.01)}	 & \textbf{0.87 (0.04)}	 & \textbf{0.91 (0.02)}\\
\bottomrule
\end{tabular}
}

\end{table}

\begin{table}[t]
\centering
\caption{Accuracy with \textit{frequently}-imputed values.}\label{tab:high_res}
\vspace{-2mm}
\resizebox{\linewidth}{!}{
\begin{tabular}{lcccc}
\toprule
\multirow{2.25}{*}{\textbf{Methods}} &  \multicolumn{4}{c}{\textbf{Datasets}}\\
\cmidrule{2-5}
    & \texttt{Walking}
    & \texttt{Running}
    & \texttt{Lying Down}
    & \texttt{Sleeping}\\      
\midrule
GRU-interp & 0.62 (0.04) 	 & 0.52 (0.05) 	 & 0.78 (0.07) 	 & 0.76 (0.05) \\
GRU-mean & 0.59 (0.02) 	 & 0.51 (0.09) 	 & 0.8 (0.04) 	 & 0.79 (0.02) \\
GRU-$\Delta$t & 0.56 (0.02) 	 & 0.48 (0.02) 	 & 0.83 (0.04) 	 & 0.80 (0.03) \\
GRU-S & 0.61 (0.04) 	 & 0.51 (0.07) 	 & \textbf{0.89 (0.02)} 	 & 0.80 (0.02)\\ 
GRU-Decay & 0.59 (0.02) 	 & 0.51 (0.05) 	 & 0.88 (0.05) 	 & 0.77 (0.02) \\
GRU-D & 0.63 (0.02) 	 & 0.45 (0.01) 	 & 0.82 (0.03) 	 & 0.73 (0.01) \\
IPN & 0.62 (0.01) 	 & 0.46 (0.01) 	 & 0.85 (0.05) 	 & 0.78 (0.04) \\
mTAN & 0.64 (0.01) 	 & 0.55 (0.01) 	 & 0.74 (0.02) 	 & 0.76 (0.02) \\
NCDE & 0.66 (0.00) 	& 0.53 (0.00) 	& 0.75 (0.00) 	& 0.71 (0.00) \\
\midrule
CAT w/o Moment & 0.61 (0.01) 	 & 0.47 (0.00) 	 & 0.77 (0.00) 	 & 0.74 (0.01) \\
CAT (ours) & \textbf{0.83 (0.03)}	 & \textbf{0.65 (0.04)} 	 & \textbf{0.89 (0.03)} 	 & \textbf{0.90 (0.01)}\\
\bottomrule
\end{tabular}
}
\vspace{-7mm}
\end{table}

Our results for this experiment, shown in Tables \ref{tab:low_res} and \ref{tab:high_res}, show that, as expected, CAT outperforms all compared methods in both the \textit{infrequent} and the \textit{frequent} settings for all datasets by an average of over 8\%.
The baselines also mainly perform their best with \textit{infrequent} imputation, while CAT performs its best at \textit{frequent} imputation as it adapts to different resolutions.
Also as expected, the recent \textit{GRU-D}, \textit{IPN}, and \textit{mTAN} models are generally CAT's strongest competitors.
As expected, methods that model irregularity (\textit{GRU-D}, \textit{IPN}, \textit{GRU-S}, and CAT) largely beat the methods that disregard irregularity.
\textit{GRU-interp}'s poor performance indicates that the benefits of CAT do not come from the linear interpolation used by the Receptor Network.


For all datasets, CAT outperforms \textit{CAT w/o Moment}, the \textit{policy-free} version of CAT that places the \textit{Receptor Network} at random moments in the timeline.
In fact, \textit{CAT w/o Moment} is overall the \textit{worst}-performing method, indicating that CAT's strong performance comes from a successfully-trained \textit{Moment Network}.
Therefore CAT indeed succeeds to \textit{learn} the discriminative regions of the given time series.
However, it is possible that \textit{CAT w/o Moment} could still perform well with enough moments.
To determine if this is the case, we vary the number of moments for both CAT and \textit{CAT w/o Moment}. 
As our results in Figure 5 in the Appendix show, the moment network is effective.



\subsubsection{Experiments on Long Synthetic Data.}\label{sec:synth_exp}
\begin{figure}[t]
    \begin{subfigure}[t]{0.95\linewidth}
        \centering
        \includegraphics[width=.99\textwidth]{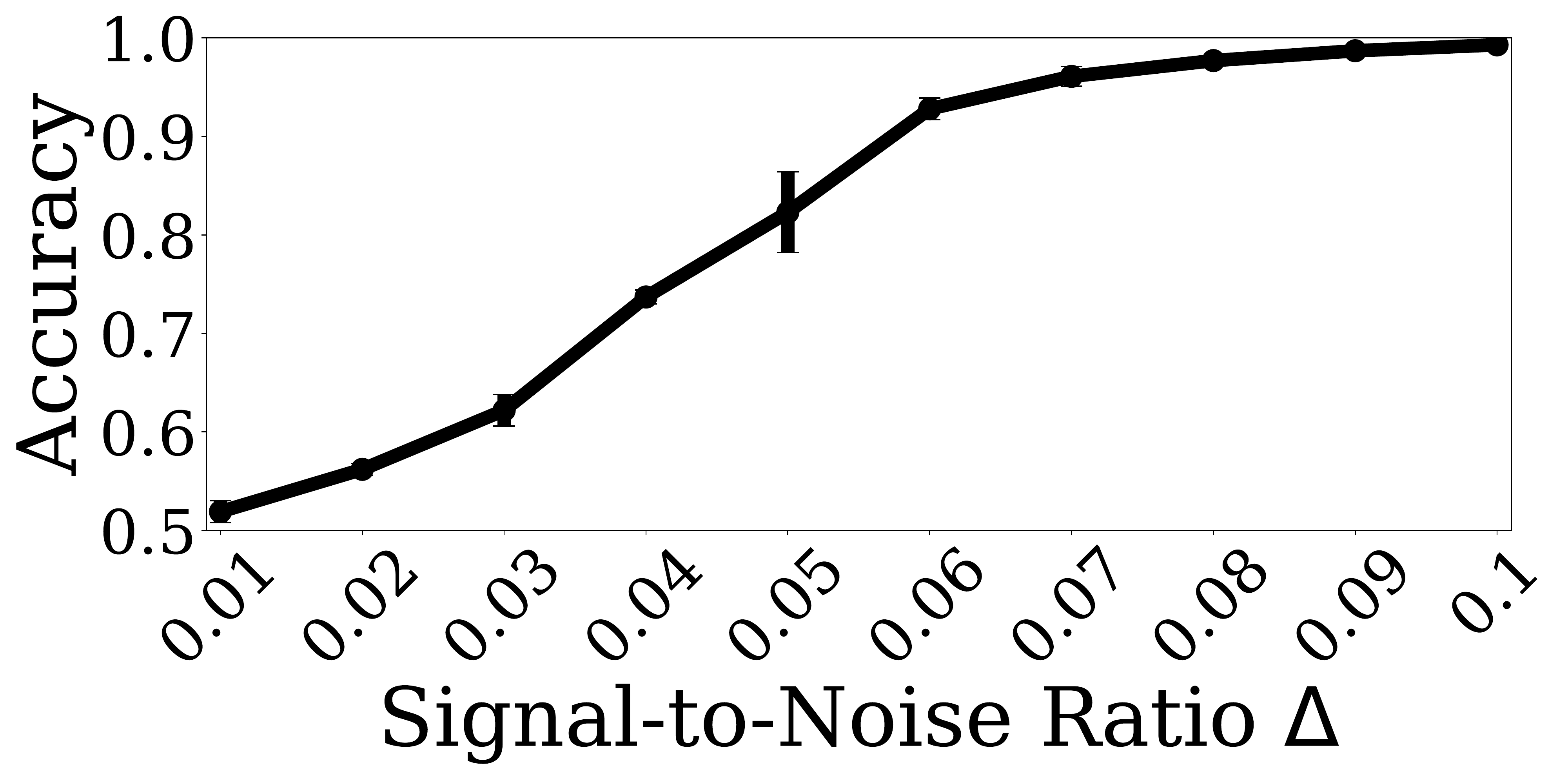}
        \caption{Effect of signal width $\Delta$.}
        \label{fig:signal_to_noise}
    \end{subfigure}
    \begin{subfigure}[t]{0.95\linewidth}
        \centering
        \includegraphics[width=.99\textwidth]{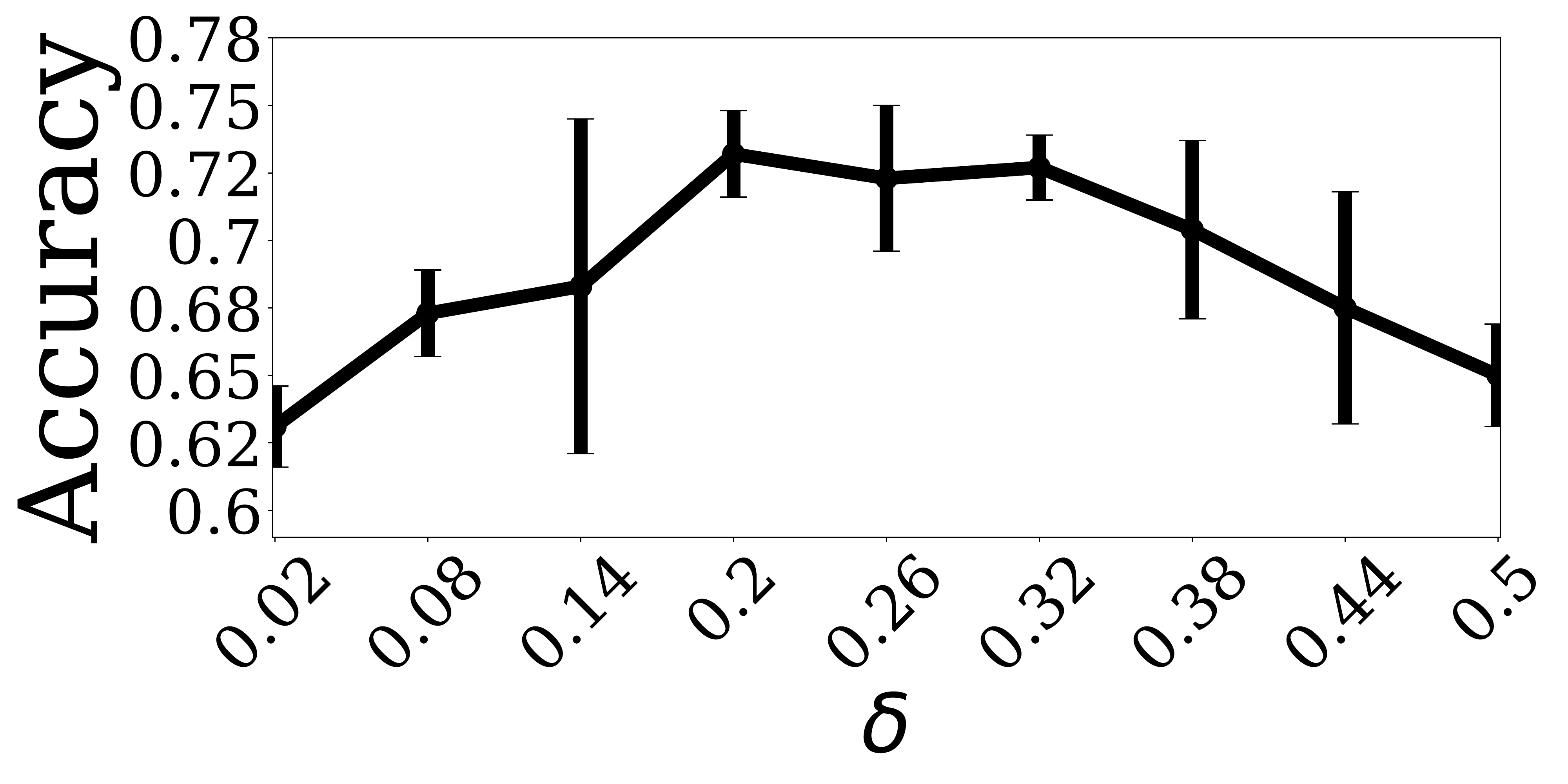}
        \caption{Effect of $\delta$ with $\Delta=0.04$.}
        \label{fig:hyperparam}
    \end{subfigure}
    \vspace{-2mm}
    \caption{CAT's performance on Synthetic $\text{M}\Pi$ dataset.}
    \label{fig:mpi}
    \vspace{-5mm}
\end{figure}

We finally evaluate CAT's robustness to signal length using the synthetic \textsc{M}$\Pi$ dataset.
We use long, 500 timestep time series for all experiments.
Therefore, for short signals, there is a huge amount of noise with very tiny relevant regions.
Our results are shown in Figure \ref{fig:mpi}.

First, as shown in Figure \ref{fig:signal_to_noise}, we vary the signal-to-noise ratio in $\text{M}\Pi$, as defined by the length of the length of the relevant signal for each class.
Intuitively, as this ratio increases, the signal becomes easier to identify.
By updating the \textit{receptor width} $\delta$ to match the signal-to-noise ratio as it is increases, we find that the \textit{Moment Network} indeed succeeds in finding the discriminative moments in the timeline, achieving nearly-perfect accuracy even when the signal only takes up 6\% of the timeline.
Once the signal takes up 10\% of the timeline, CAT consistently achieves 100\% testing accuracy.
We also find that the compared methods fail when the signal-to-noise ratio is lower than 0.1, achieving roughly 50\% testing accuracy.
This is expected as RNNs are classically hard to train on long series, especially with such noisy inputs.

Second, as shown in Figure \ref{fig:hyperparam}, we vary the \textit{receptor width} parameter $\delta$ for a signal-to-noise ratio $\Delta$ of 0.04 to understand CAT's sensitivity to the proper selection of $\delta$.
We investigate the signal-to-noise ratio of 0.04 where CAT achieves only 75\% accuracy, indicating potential sensitivity to hyperparameters (see Figure \ref{fig:signal_to_noise}).
As expected, accuracy suffers both when $\delta$ is either too small (0.02) or too large (0.5).
The optimal $\delta$ lies somewhere between 0.2 and 0.32 for this experiment.
Quite interestingly, this is much larger than the data's signal-to-noise (0.04).
While a larger receptor width $\delta$ should capture signals more easily, suggesting that the receptor still filters out the noisy regions when they overlap with the receptor's window.
These results also indicate that CAT can be robust to overestimating $\delta$. 

\section{Conclusions}
\label{sec:conclusions}
In this work, we identify the open Attention-Based Classification problem for long and irregularly-sampled time series, which is a challenging and impactful setting common to many important domains.
The Attention-Based Classification problem is to classify long irregularly-sampled time series based on small discriminative signals in the continuous timeline while learning to ignore irrelevant regions.
Since prior methods rely on good selection of a set of timesteps at which to impute values, they struggle to classify time series in this setting, which we demonstrate experimentally.
Using insights from prior methods, we then propose the Continuous-time Attention Policy Network (CAT), which generalizes previous works by learning to searching for short signals in a time series' potentially-long timeline.
CAT includes a reinforcement learning-based Moment Network that seeks discriminative moments in the timeline, positioning a novel Receptor Network that represents signals from \textit{both} the values themselves and the patterns existing in the timing of the observations.
Using a core Transition Model that learns to model the transition between moments, a Discriminator Network finally classifies the entire series.
This approach can intuitively be extended to match the modeling paradigms proposed by other recent methods, like differential equation models and time-representational encodings.
We validate our method on a wide range of experiments featuring four real datasets, ablation studies, impacts of hyperparameter selection, a synthetic data highlighting CAT's strengths, and timing experiments.
Across the board, CAT consistently outperforms recent alternatives by successfully finding short signals in long time series.



\balance
\bibliography{main.bib}
\bibliographystyle{icml2022}

\newpage
\appendix
\section*{Appendix}

\section{Dataset Descriptions}\label{sec:har}
All datasets statistics are shown in Table \ref{tab:datastats}. For each of the four Human Activity Recognition datasets from the \textsc{ExtraSensory} dataset (\texttt{http://extrasensory.ucsd.edu/}) \cite{vaizman2017recognizing} (\textsc{Walking}, \textsc{Running}, \textsc{LyingDown}, and \textsc{Sleeping}), we aim for an 80\% training and 20\% testing split \textit{in time}, though this is challenging to control in practice. Thus, the exact ratio differs between the series.
For each dataset, we further split off 10\% of the training set for validation.

\begin{table}[h]
  \caption{Dataset Statistics. $\text{N}_\text{train}$ denotes the number of training instances, $\text{N}_\text{train}$ is the number of testing instances, Avg. $T$ is the average number of observations per series, and $C$ is the number of classes.}
  \centering
  \label{tab:datastats}
    \begin{tabular}{lcccc}
    \toprule
    Dataset & $\text{N}_\text{train}$ & $\text{N}_\text{test}$ & Avg. $T$ & $C$\\
    \midrule
    \textsc{$\text{M}\Pi$} & 4000 & 1000 & 500 & 2\\
    \textsc{UWave} & 4030 & 448 & 94 & 8\\
    \textsc{Walking} & 1616 & 1020 & 99 & 2\\
    \textsc{Running} & 666 & 400 & 85 & 2\\
    \textsc{LyingDown} & 6186 & 1240 & 80 & 2\\
    \textsc{Sleeping} & 6462 & 2814 & 80 & 2\\
  \bottomrule
\end{tabular}
\end{table}

\section{Further M$\Pi$ Experiments}\label{sec:app_synth}
Expanding on the synthetic experiment discussed in the Experiments section of our main paper, we also run all compared methods for each of the signal-to-noise ratios, the results of which are shown in Figure \ref{fig:synth_results_all}. Again, each series has 500 timesteps, only 3 of which are relevant to the classification task. The 3 relevant timesteps are evenly-spaced in a randomly-placed width-$\Delta$ window in the continuous timeline. In this experiment, all compared methods fail to classify these series, even when performing imputation with 500 timesteps, which does not delete the signal. Instead, they fail to \textit{focus} on the the discriminative region and so cannot perform classification. On the contrary, CAT achieves nearly-perfect accuracy with a signal-to-noise ratio as low as .06, indicating that it indeed does find the relevant regions.

\section{Timing Experiments}\label{sec:timing}
CAT's \textit{Transition Model} uses an RNN to model the transitions between \textit{moments}, as opposed to the timestamps themselves.
This hints that CAT should naturally be much faster than the compared methods.
We confirm this by timing the training of all methods on the \textsc{Walking} dataset with frequent imputation---see Figure \ref{fig:timing}.
As expected, CAT runs over seven times faster than the next slowest method while achieving much higher testing accuracy.
This is particularly meaningful for long series in time-sensitive domains such as healthcare where a model's inference time is hugely important \cite{hong2020holmes}.
Our reported timing comparisons between compared methods is also largely consistent with prior works' timing experiments \cite{shukla2019interpolation}.
Also as expected, the GRU-D \cite{che2018recurrent}, mTAN \cite{shukla2021multi}, and NCDE \cite{kidger2020neural} run significantly slower than the other compared methods and so omit their results from this figure.
Their Accuracies are much lower than CAT's---see Table 2 in the main paper.
All models were trained and evaluated on Intel Xeon Gold 6148 CPUs.

\begin{figure}[htp]
    \centering
    \includegraphics[width=0.44\textwidth]{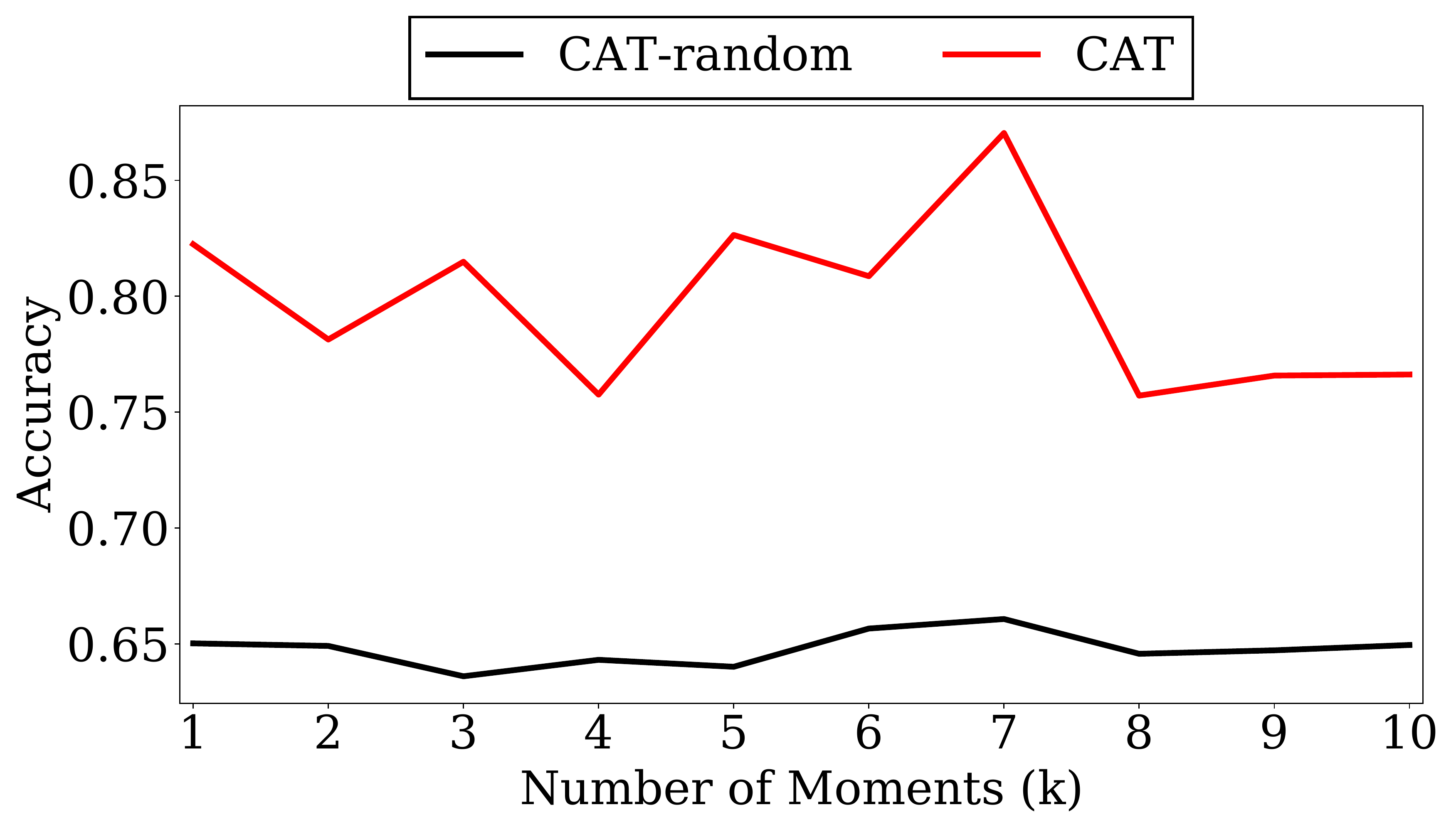}
    \vspace{-5mm}
    \caption{Ablating the impact of the Moment Network using the \textsc{Walking} dataset. CAT outperforms CAT-random so the number of hops does not increase accuracy alone.}
    \label{fig:nhop}
\end{figure}

\begin{figure}[h]
    \centering
    \includegraphics[width=0.8\linewidth]{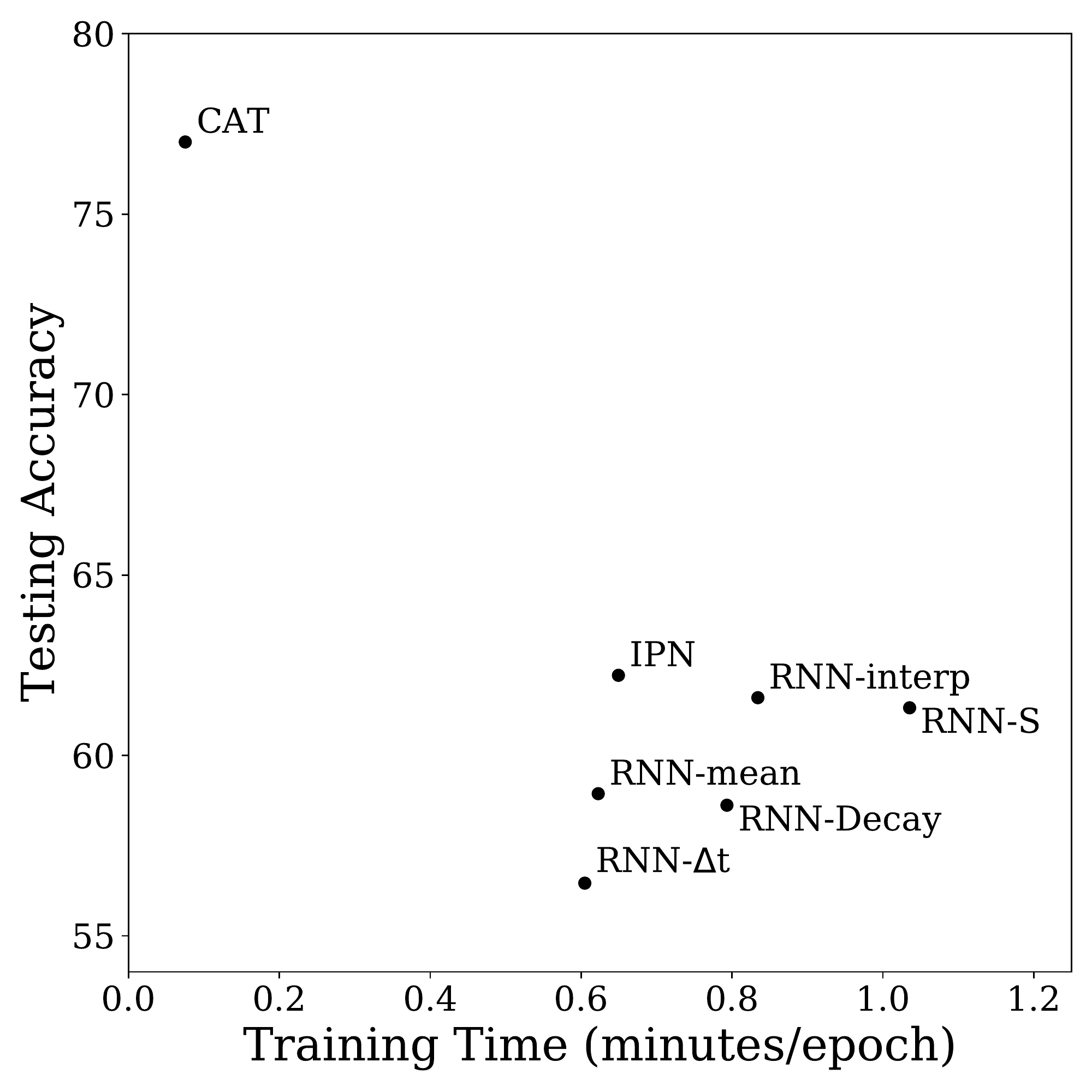}
    \vspace{-3mm}
    \caption{Timing performance for the high-resolution \textsc{Walking} dataset. GRU-D and NCDE take over 3x longer than the next-slowest RNN-S and so is omitted from this figure.}
    \label{fig:timing}
\end{figure}

\section{CAT Hyperparameters}\label{sec:hyperparam}
We experiment with three key hyperparameters of CAT for each dataset: The receptor-width $\delta$, the hidden dimension of the Receptor Network $\mathcal{R}$, and whether or not to use the informative irregularity feature of CAT in the Receptor Network.
Interestingly, we found that for $\mathcal{R}$, a hidden dimension of 50 seemed to consistently produce the best results.
This hidden dimension largely controls the number of parameters in CAT and influences the timing experiments for which we also use a 50-dimensional representation.
Our selections for $\delta$ values for different datasets are shown in Table \ref{tab:hyperparam}.

We tune $\delta$ between three values: $0.05$, $0.1$, and $0.2$. 
For \textsc{UWave}, $\delta=0.05$ was best.
$\delta=0.05$ was also best for all infrequent \textsc{ExtraSensory} datasets except for \textsc{WALKING}, which used $0.2$.
$\delta=0.2$ was chosen for all frequent \textsc{ExtraSensory} datasets except for \textsc{SLEEPING}, for which $\delta=0.1$.

For $\delta$, we observe that for the \textit{infrequent} experiments, a smaller receptor width is largely the best option while a larger width is beneficial for the \textit{frequent} experiments. This may be due to the fact that with the infrequent representation of the input series, closer focus on the comparatively-fuzzier signals is required.
We also found that setting the number of steps $k=3$ consistently outperformed larger and smaller values.
While large values of $k$ conceptually should still learn to classify effectively, in practice the more steps taken by a reinforcement learning agent per episode can make it more challenging to optimize effectively due to the credit assignment problem. 

We also find that there are cases where it is not essential to use both channels---Values and Irregularity---in the receptor network.
While using the irregularity channel (computed via the squared exponential kernel in the main paper) always leads to state-of-the-art performance by CAT, its omission can sometimes improve CAT's performance slightly.
When irregularity is an essential feature, however, this information cannot be removed.
We show for which datasets this is true in Table \ref{tab:hyperparam}.
This may be a feature of (1) how the irregularity is represented---there are other approaches---and (2) how essential it is to the task.
We recommend always using the irregularity channel as the potential downside of ignoring irregularity outweighs the minor benefits of omission in some cases.

\begin{table}[t]
  \caption{Best hyperparameter settings for CAT.}
  \resizebox{\linewidth}{!}{
  \centering
  \label{tab:hyperparam}
    \begin{tabular}{lcccc}
    \toprule
    Dataset & $\delta$ &  Hidden Dim. of $\mathcal{R}$ & Density\\
    \midrule
    \textsc{UWave} & 0.05 & 50 & Off  \\
    Infrequent \textsc{Walking} & 0.2 & 50 & Off \\
    Infrequent \textsc{Running} & 0.05 & 50 & On \\
    Infrequent \textsc{LyingDown} & 0.05 & 50 & Off \\
    Infrequent \textsc{Sleeping} & 0.05 & 50 & On \\
    Frequent \textsc{Walking} & 0.2 & 50 & Off \\
    Frequent \textsc{Running} & 0.2 & 50 & On \\
    Frequent \textsc{LyingDown} & 0.2 & 50 & On \\
    Frequent \textsc{Sleeping} & 0.1 & 50 & Off \\
  \bottomrule
\end{tabular}
}
\end{table}

\begin{figure}[h]
    \centering
    \includegraphics[width=0.99\linewidth]{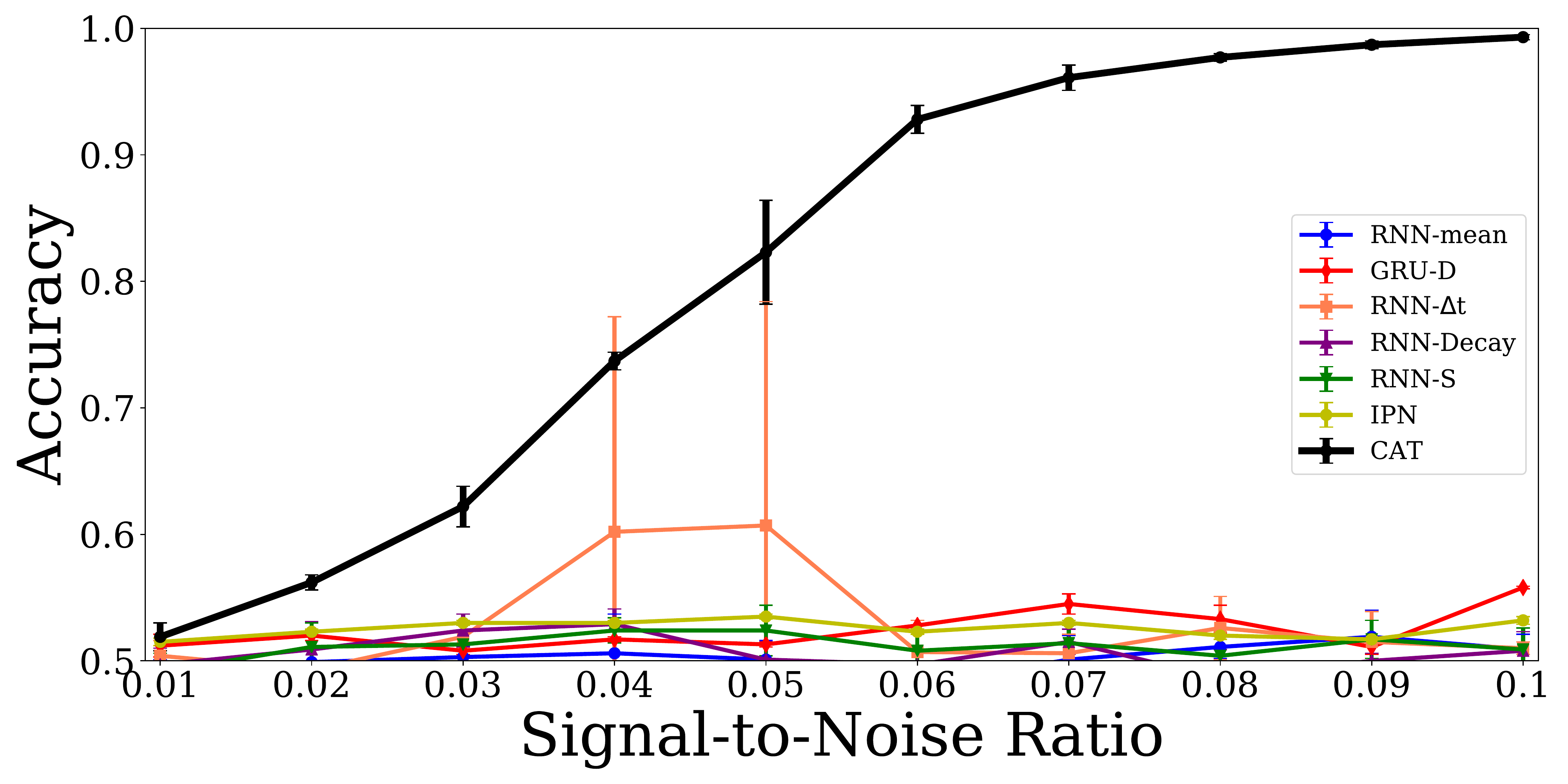}
    \caption{Effect of changing the signal width on accuracy.}
    \label{fig:synth_results_all}
\end{figure}



\end{document}